# Artificial Retina Using A Hybrid Neural Network With Spatial Transform Capability

Richard Wood
Honeywell, Inc., Space Payloads
Ontario, Canada

Alexander McGlashan
Niagara College
Welland, Ontario

C.B. Moon
Hoseo University
South Korea

W.Y. Kim
Hoseo University
South Korea

*Abstract*—This paper covers the design and programming of a hybrid (digital/analog) neural network to function as an artificial retina with the ability to perform a spatial discrete cosine transform. We describe the structure of the circuit, which uses an analog cell that is interlinked using a programmable digital array. The paper is broken into three main parts. First, we present the results of a Matlab simulation. Then we show the circuit simulation in Spice. This is followed by a demonstration of the practical device. This system has intentionally separated components with the specialty analog circuits being separated from the readily available digital field programmable gate array (FPGA) components. Further development includes the use of rapid manufacture-able organic electronics used for the analog components. The planned uses for this platform include crowd development of software that uses the underlying pulse based processing. The development package will include simulators in the form of Matlab and Spice type software platforms.

## I. Introduction

This paper covers the design and programming of a hybrid (digital/analog) neural network to function as an artificial retina with the ability to perform a spatial discrete cosine transform. We describe the structure of the circuit, which uses an analog cell that is interlinked using a programmable digital array. The paper is broken into three main parts. First, we present the results of a Matlab simulation. Then we show the circuit simulation in Spice. This is followed by a demonstration of the practical device. This system has intentionally separated components with the specialty analog circuits being separated from the readily available digital field programmable gate array (FPGA) components. Further development includes the use of rapid manufacture-able organic electronics used for the analog components. The planned uses for this platform include crowd development of software that uses the underlying pulse based processing. The development package will include simulators in the form of Matlab and Spice type software platforms.

## II. Background

There are a number of hybrid neural network designs in development[1-80]. These include Spinnaker, True north. Neurogrid, and Brainscales. Spinnaker and True North use an Address Event Representation (AER). In Neurogrid, analog neural effects are modelled and synaptic paths are shared. The current system uses an approach in which all synaptic connections exist at all times in a high speed serial architecture.

The discrete cosine spatial transform can be used for texture analysis. The variations in texture in objects can be used to determine relative distance as well as defining the structure itself. These cosine transforms demonstrate one aspect of the circuit. By definition, any function is programmable including complex functions. The circuit can also be dynamically reprogrammed for different functions while in operation.

## III. Basic Device Description

The following is an overview of the hybrid (analog-digital) design of the neural processor. There are two main points to note. 1. The design involves the use of analog neural network cells that are interlinked using an integrated digital interface. 2. The base design described here is an 8x8x 3 layer hybrid neural net, specifically designed for use as a texture analyzing circuit based on a discrete cosine transform algorithm.

The cells can be programmed through a scanning array via the digital interface. The digital interface in this implementation is composed of an FPGA which has be designed in two main configurations. One with a high level of parallel processing and the other in a high speed serial configuration. The analog cells are monitored through a scanning array and the information is then digitized and stored in the FPGA. The scanning occurs at least 1000 times the rate of pulses (MHz versus kHz). The scanned cell information is then stored in a register. This register is used to calculate the total excite or inhibit synaptic input for each cell of the next layer. The input cell register is scanned and the weighted effect for each synaptic connection is stored in an accumulator for the designated receptor cell in the next layer. These accumulators are structured according to delay (complex weight) so that they are stored in an accumulator that is used during a future cycle. The weights can be dynamically updated to change the system interlinking configuration. The information is then converted into resultant "time-on" information to be passed to the connected synapse. The resultant "time-on" will be either inhibitory or excitatory. When the individual cell has been on long enough the input current will be turned off. This system works on the premise that the average cell input current based on total on time is equivalent to the summation of input inhibitory and excitatory pulses. It is time scale dependent. The current configuration uses pulse rate information for processing. An arrival time system is possible within the scanning and processing time limits imposed by the system.

In the example of the 8x8x3 layer system, the first layer is an array of detector cells, converting light intensity into an array of pulse rates. This is then subject to a 1D row discrete cosine transform, with the information still expressed as pulse frequencies. This second 8x8 array is then also subject to a 1D transform (along the columns). The result is a 2D cosine transform encoded as variable rate pulses. The transform is a direct result of the stored weights.

Figure 1 shows an illustration of the 8x8x3 layer neural network system. Each circle represents an analog pulsing integrating cell. These cells are interconnected via a raster system and bus lines that feed into the FPGA and processor, as in Figure 2. The first layer is continuously scanned for output pulses. When a pulse is detected, it is stored in a transient look-up table tabulated by source location. There are 64 individual source locations in the first layer. These are then converted into a following table by multiplying by stored weight values and accumulating according to destination cell and time delays. The delay size and resolution can be altered as required. This information is then encoded into a total time on for the destination cell (either an inhibitory or excitatory connection). This effectively means that although each cell in layer 2 has 128 synaptic input connections and 128 output connections, there is only one each of inhibit and excite physical connections for a total of 4 connections per cell. There are 128 virtual inhibit and excite synapses in the digital domain. The analog cells can also be programmed (threshold voltage, refractory time, etc.) via the integrated interface.

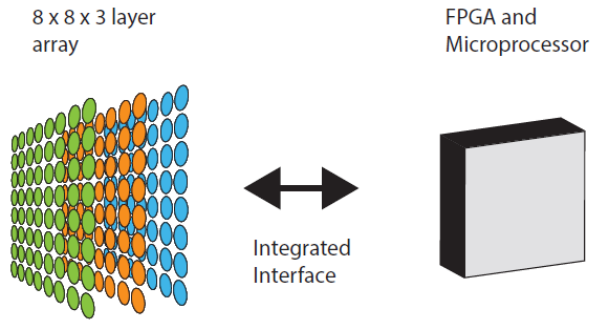

Figure 1.　System Overview

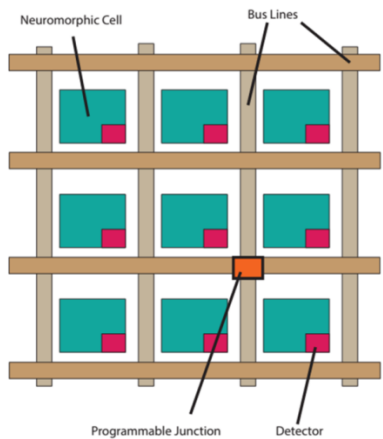

Figure 2.　Sample of Array

Figure 3 shows a sample of the neuromorphic analog cell. Figure 4 show a typical membrane potential of the analog cell.

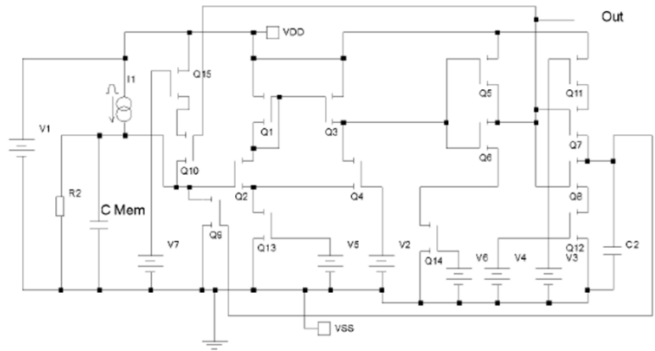

Figure 3. Analog Neuromorphic cell.

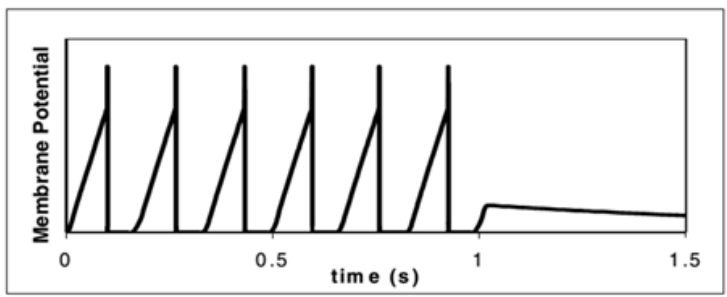

Figure 4. Typical Membrane Potential of Neuromorphic cell.

Figure 5 shows the scanning array of neuromorphic cells. At high speed, the analog cells are scanned for pulses. The scan time is orders of magnitude less than the maximum pulse rate.

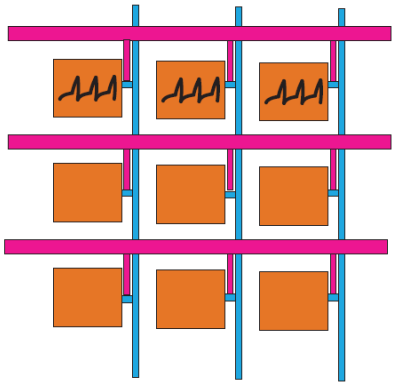

Figure 5. Scanning Array of Neuromorphic cells.

Figure 6 and 7 illustrate the weight calculation. Each of the neurons in the first layer is connected to every cell in the following layer by a given weight which includes a time delay (complex connection). A destination-delay table is used to store and accumulate these weighted results. For example, if neuron A, B, C and D fired, they are then connected to neuron 1 in the next layer. But, we assume that the weight function between C and 1 has a delay different than A, B and D. In this case, the weighted results of A, B and D are accumulated and stored in one column of the table and the weighted result of C is stored in a delayed column for neuron 1. This table is constantly upgraded and shifted according to the time delay. The resolution and range of the time delay determines the size of the table. Since the weights can be either excitatory or inhibitory, the resultant totals are either one or the other as well.

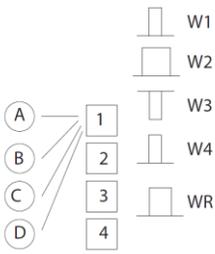

Figure 6.  Weight Calculation.

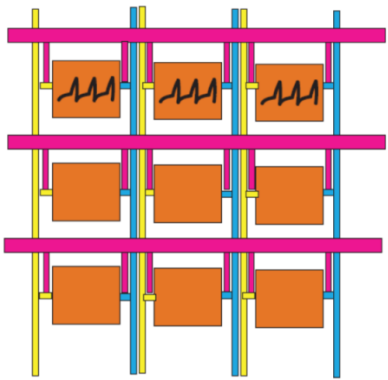

Figure 7.  Destination-Delay Table.

Figure 8 shows the synaptic stimulating scanning array. The data in the destination-delay table, in column 1, is now interpreted as resultant "time-on" (either inhibit or excite). The array shown is scanned by column, any neurons in that column that have a total current on value are simultaneously turned on (excite or inhibit). The next column is then scanned. Once all columns are scanned, the system goes back to the first and turns off any neuron currents that have reached their total "on-time" value. The system effectively digitally integrates pulse trains over short periods of time and converts them into total current values for that short time period as shown in figure 9.

Figure 8.  Scanning input pulse array.

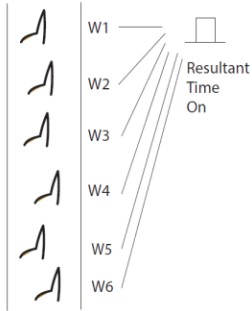

Figure 9. Pulse-Time Conversion

## IV. DISCRETE COSINE TRANSFORM

For an 8-by-8 matrix of neural cells, T*A is an 8-by-8 matrix whose columns contain the one-dimensional DCT of the columns of neural cell matrix A. The two-dimensional DCT of the neural matrix A can be computed as B=T*A*T', with B being the output matrix and T and T' are the weight matrix and its transpose.

## V. DISCUSSION

### A. Matlab simulation

In Matlab we designed the simulation as per the system illustrated in Figure 1. The simulation uses an analog cell approach with a runge kutta solution. The weights are programmed discretely. The inputs to the first layer are simulated as relative light intensity. Figure 10A illustrates the input 8 x 8 array, showing the layer output of a single sample test pattern. Figure 10B illustrates the hidden 8 x 8 array layer that performs the first one dimensional DCT. Figure 10C illustrates the 8 x 8 output array layer that performs the second one dimensional DCT. A complete series of test patterns was evaluated.

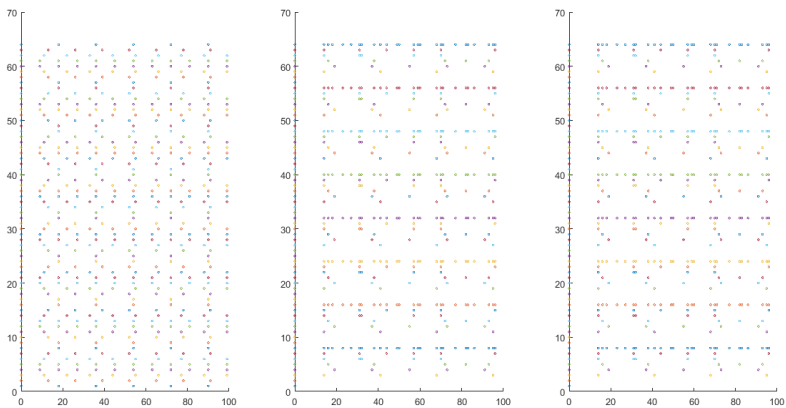

Figure 10. A) Input Layer Pulse Graph B) Hidden Layer Pulse Graph C) Output Layer Pulse Graph

## B. Spice Simulation

The following section gives a high level overview of the Spice simulation. The system comprises a grid of neural cells as shown in Figure 11, with a representative individual cell shown in Figure 12. The control circuit (FPGA) is implemented digitally (not shown). Figure 13 illustrates the representative integrated input to a single neuron with the neurons membrane potential.

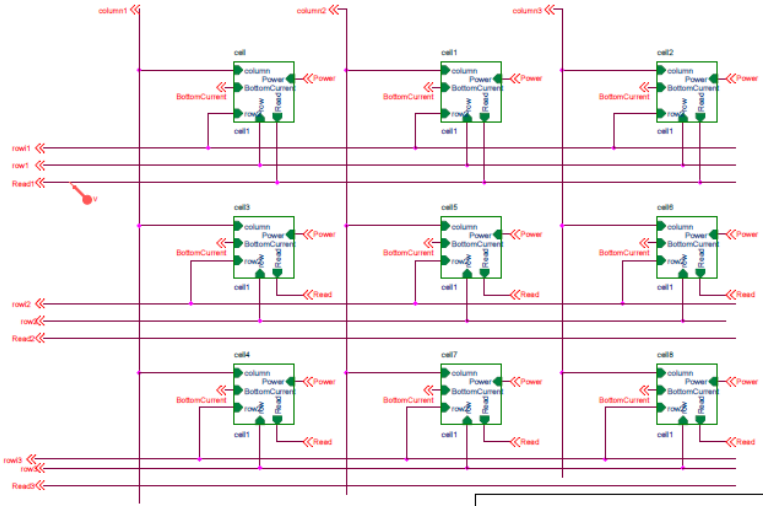

Figure 11. Neural Array (Selection 3 x 3)

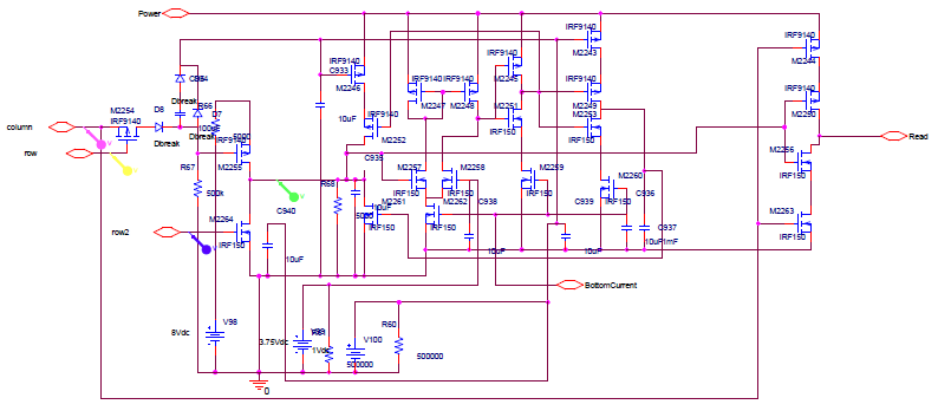

Figure 12. Basic Cell

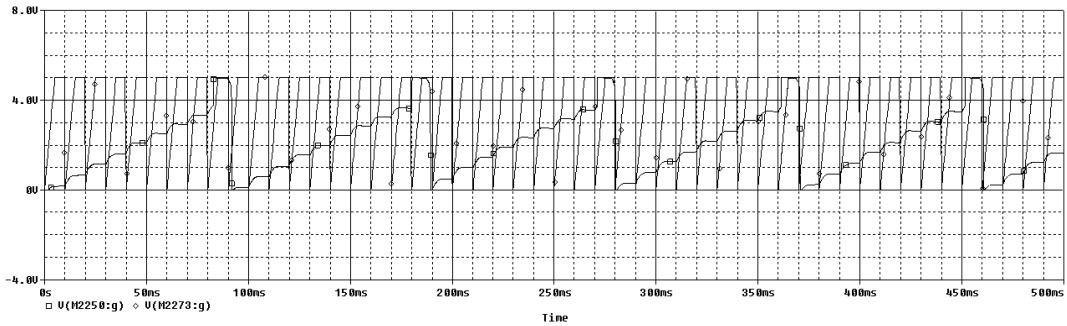

Figure 13. Representative pulsing input of a single neuron.

## C. Practical Device.

The practical device was implemented in two forms. 1. A hybrid device with the control circuitry designed in an Altera FPGA, with the digital processor combined with external analog circuits (neural cells), in a scanning matrix. 2. A system in which the entire design is implemented in an Altera FPGA with the analog neural cells and the scanning array emulated within the FPGA.

### 1) Hybrid Implementation

Figure 14 illustrates the functional schematic of the hybrid system. This design was implemented in two versions, a parallel and a serial configuration. The primary difference being that in the parallel version, each destination cell calculation is done in parallel, requiring 64 separate circuits.

Referring to Figure 14, the calculate outputs block defines the time sequencing. The weights are loaded from RAM. The accumulator for each destination neural cell is updated in sequence as each of the 64 input cells firing register is scanned for activity and the designated weight is added to the accumulator of the destination cell. These calculations result in a requirement for 4096 cycles to calculate all the output amounts in the serial version and 64 cycles in the parallel version. In this example, that results in a 82 us cycle time for the serial version.

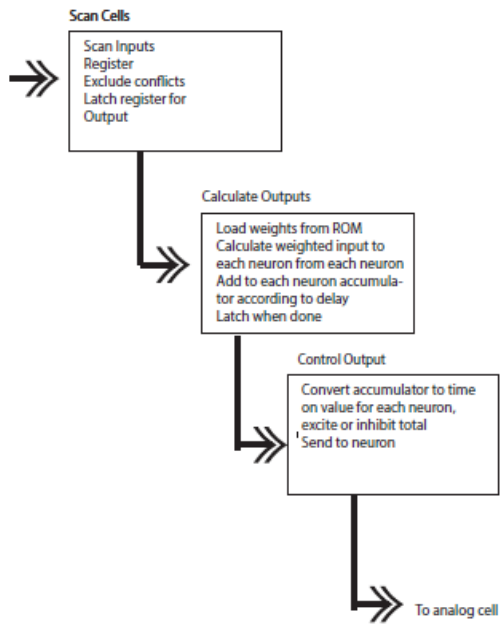

Figure 14. Functional Schematic of the system with external analog neuron.

Table 1 illustrates the FPGA resource usage for the design with an external analog neural cell. Shown are the two versions, one with the implementation using mostly logical elements in parallel and a serial version using mostly on-chip RAM. In either case, the resource usage will scale as the square of the number of neurons. In the case of the latter implementation, the logical element usage is 3 per neuron.

| Table 1. FPGA with External Analog Cell | | | |
|---|---|---|---|
| Parallel Version | | Serial Version | |
| Flow Status | Successful - Tue Mar 27 20:13:15 2018 | Flow Status | Successful - Thu Apr 05 18:23:42 2018 |
| Quartus II 64-Bit Version | 13.0.0 Build 156 04/24/2013 SJ Web Edition | Quartus II 64-Bit Version | 13.0.0 Build 156 04/24/2013 SJ Web Edition |
| Revision Name | DE2_D5M | Revision Name | DE2_D5M |
| Top-level Entity Name | NeuralNetwork2 | Top-level Entity Name | NeuralNetwork2 |
| Family | Cyclone II | Family | Cyclone II |
| Device | EP2C35F672C6 | Device | EP2C35F672C6 |
| Timing Models | Final | Timing Models | Final |
| Total logic elements | 5,910 / 33,216 ( 18 % ) | Total logic elements | 575 / 33,216 ( 2 % ) |
| Total combinational functions | 5,588 / 33,216 ( 17 % ) | Total combinational functions | 535 / 33,216 ( 2 % ) |
| Dedicated logic registers | 3,244 / 33,216 ( 10 % ) | Dedicated logic registers | 305 / 33,216 ( < 1 % ) |
| Total registers | 3244 | Total registers | 305 |
| Total pins | 258 / 475 ( 54 % ) | Total pins | 258 / 475 ( 54 % ) |
| Total virtual pins | 0 | Total virtual pins | 0 |
| Total memory bits | 49,152 / 483,840 ( 10 % ) | Total memory bits | 35,840 / 483,840 ( 7 % ) |
| Embedded Multiplier 9-bit elements | 0 / 70 ( 0 % ) | Embedded Multiplier 9-bit elements | 0 / 70 ( 0 % ) |
| Total PLLs | 0 / 4 ( 0 % ) | Total PLLs | 0 / 4 ( 0 % ) |

Figure 15 shows the chip resource usage for the serial version with the external analog neural cell.

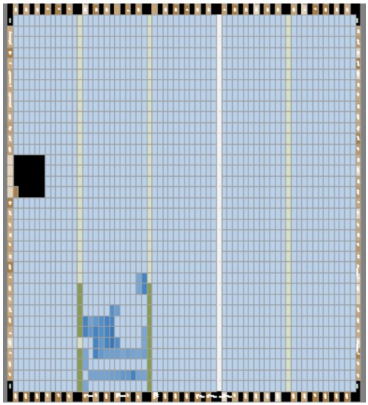

Figure 15. Chip Usage, Serial Version with External Analog Neural Cell.

*2) Full FPGA Implementation*

Figure 16 illustrates the functional schematic of the system in which the entire design is implemented in an Altera FPGA with the analog neural cells and the scanning array emulated within the FPGA.

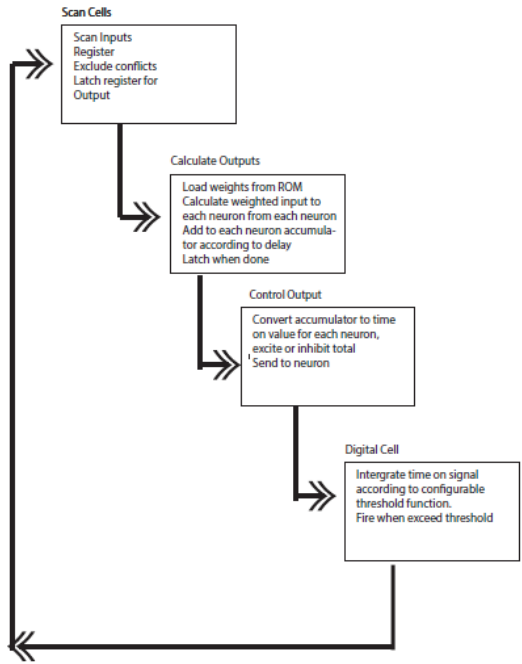

Figure 16. Implementation with Neural Cell Emulated in The FPGA.

Table 2 illustrates the FPGA resource usage for the design with the neural cell emulated within the FPGA. Shown are two versions, one with the implementation using mostly logical elements in parallel and a serial implementation using mostly on chip RAM. Comparing to the version with the external analog neural cell, the neuron requires 9 logical elements to emulate each neuron. As resources other than the neural cell scale by the square of the number of neurons and the cell resource usage scales linearly with the number of neurons, the resource usage advantage of one implementation over the other is design specific.

| Table 2. FPGA with Emulated Neural Cells. | | | |
|---|---|---|---|
| Parallel Version | | Serial Version | |
| Flow Status | Successful - Thu Mar 29 17:41:43 2018 | Flow Status | Successful - Sat Apr 07 09:21:00 2018 |
| Quartus II 64-Bit Version | 13.0.0 Build 156 04/24/2013 SJ Web Edition | Quartus II 64-Bit Version | 13.0.0 Build 156 04/24/2013 SJ Web Edition |
| Revision Name | DE2_D5M | Revision Name | DE2_D5M |
| Top-level Entity Name | NeuralNetwork2 | Top-level Entity Name | NeuralNetwork2 |
| Family | Cyclone II | Family | Cyclone II |
| Device | EP2C35F672C6 | Device | EP2C35F672C6 |
| Timing Models | Final | Timing Models | Final |
| Total logic elements | 11,155 / 33,216 ( 34 % ) | Total logic elements | 1,043 / 33,216 ( 3 % ) |
| Total combinational functions | 10,799 / 33,216 ( 33 % ) | Total combinational functions | 993 / 33,216 ( 3 % ) |
| Dedicated logic registers | 4,584 / 33,216 ( 14 % ) | Dedicated logic registers | 427 / 33,216 ( 1 % ) |
| Total registers | 4584 | Total registers | 427 |
| Total pins | 258 / 475 ( 54 % ) | Total pins | 258 / 475 ( 54 % ) |
| Total virtual pins | 0 | Total virtual pins | 0 |
| Total memory bits | 49,152 / 483,840 ( 10 % ) | Total memory bits | 35,968 / 483,840 ( 7 % ) |
| Embedded Multiplier 9-bit elements | 64 / 70 ( 91 % ) | Embedded Multiplier 9-bit elements | 0 / 70 ( 0 % ) |
| Total PLLs | 0 / 4 ( 0 % ) | Total PLLs | 0 / 4 ( 0 % ) |

Figure 17 shows the chip resource usage for the serial version with the external analog neural cell.

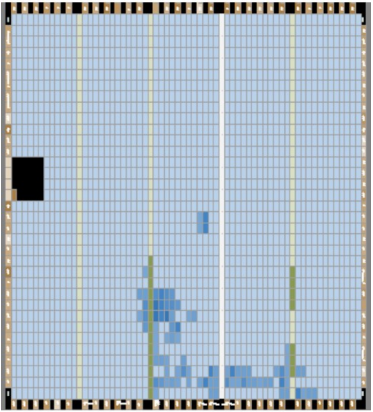

Figure 17. Chip Usage, Serial Version with Emulated Neural Cell.

## VI. Results

### A. Timing issues with time sequenced synapses.

Timing error results from the conversion of pulses into time on/off of the synapses. With the present serial implementation, there is an approximate 82 us requirement for the calculation of the weighted sums of synaptic inputs. This then defines a scan period of 82 us. A staggered system is possible but the complications were deemed to be unnecessary. This results in a total possible delay of between 82 and 246 us. In the worst case scenario, a pulse begins at the beginning of a cycle and then the current is on until the end of a cycle (maximum synaptic connection). Figure 18 shows the time spread of delays in a Monte Carlo simulation of 10000 input pattern variations.

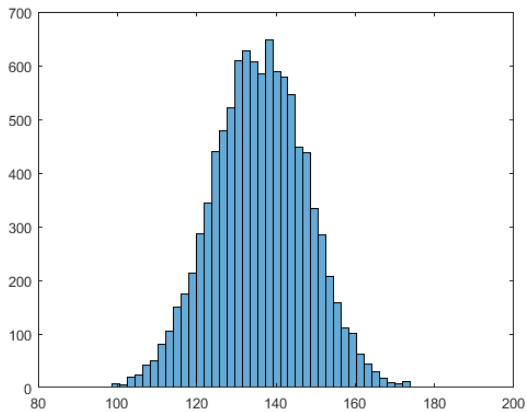

Figure 18. Time Spread Of Delays In A Monte Carlo Simulation Of 10000 Input Pattern Variations

The decay function for the analog cell is on the order of 10%/ ms. The delay will cause an error of less than or equal to 0.1%. This is well below the accuracy of the pulse train system. The quantization of the delay periods also creates an error.

### B. Stability Analysis of Digital-Analog Domain.

A basic phase state analysis follows. Figure 19 compares the quantized (dots) and non-quantized (solid line) synaptic pulses, in terms of the phase state. A sample of the quantized input is shown in Figure 13.

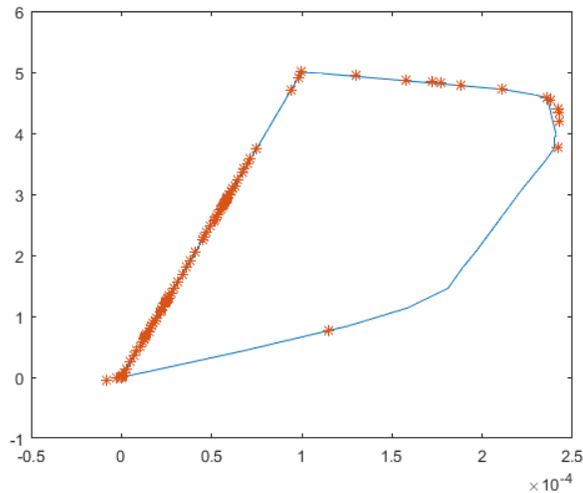

Figure 19. Phase Analysis.

### C. Learning (Digital)

Learning is mostly in the digital domain with some programmable features available in the analog implementation. The circuit can use numerous learning algorithms as it is external to the implementation. The time sequencing and pulse activity is available within the system for further enhanced learning that may include arrival time. A normalized hebbian algorithm has been implemented, although the DCT configuration is considered hard-wired.

### D. Accuracy

In biological optical neural activity there are firing rates up to 250 to 1000 Hz with averages in the 9 to 18 Hz. Using a nominal 10 Hz firing rate, if we encode only with pulse rate of a single neural train, we have a maximum accuracy of 10% (using a 1 second evaluation period). In a practical device, we have the option of adding more neurons in parallel to split the input and increase the accuracy. In a practical system, this would also reduce the effect of lost neurons. Other encoding options are possible such as the use of a power scale (logarithmic). For the present example of a spatial frequency decoder, a 10% accuracy is sufficient. If using pulse timing, the accuracy is limited to the scanning time period, 82 us giving a nominal accuracy of 0.0008%. In the context of saccades, the frequency component of the image (spatial transform) is technically independent from the effect of image jitter.

### E. Power Consumption

A detailed analysis of comparative power consumption is beyond the scope of the present work. However, the analog hybrid version uses significantly less components in the integrator.

### F. Noise

Noise scales linearly within the system. Noise of spatial frequencies that are measured by the system will be passed through to the output layer, Figure 20. Implementing a recurrent neural network will allow for temporal noise filtering.

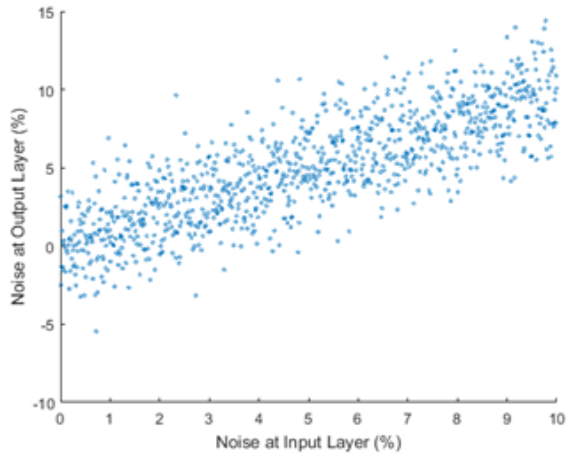

Figure 20. Noise.

*G. Accessible Tool for Students, rapid prototyping, application specific.*

Goal is to make an accessible tool. The digital-analog design simplifies rapid prototyping.

## VII. Conclusions

This work has outlined the design, implementation and testing of a pulsing hybrid neural circuit using an Altera FPGA in conjunction with an external analog neural cell matrix. An all FPGA version has also been developed as a point of comparison. The delay and accuracy reduction due to quantization has been found to be below the acceptance threshold. The phase stability is unaffected by the quantization. Future work includes the development of abstraction layers for control of the system as well as implementation in an organic electronics matrix.